\newcommand{\fig}[1]{Figure~\ref{#1}}
\newcommand{\sect}[1]{Section~\ref{#1}}
\newcommand{\tbl}[1]{Table~\ref{#1}}

\newcommand{\ignorethis}[1]{}

\newcommand{\xpar}[1]{\vspace{-3mm}\paragraph{\normalfont\bf #1}\ \ }

\newcommand{\comm}[1]{}

\newcommand{\eg}{{e.g.}\@\xspace}
\newcommand{\ie}{{i.e.}\@\xspace}
\newcommand{\etal}{{et al. }\@\xspace}

\newcommand\blfootnote[1]{%
  \begingroup
  \renewcommand\thefootnote{}\footnote{#1}%
  \addtocounter{footnote}{-1}%
  \endgroup
} 

\documentclass[runningheads]{llncs}
\usepackage{graphicx}
\usepackage{xspace}
\usepackage{amsmath,amssymb} %
\usepackage{times}
\usepackage{epsfig}
\usepackage{amsmath}
\usepackage{amssymb}
\usepackage{enumitem}
\usepackage{tabularx, booktabs}
\usepackage{caption}
\usepackage{breakcites}
\usepackage[bottom]{footmisc}

\usepackage[rightcaption]{sidecap}
\usepackage{ltablex}
\usepackage[table]{xcolor}
\definecolor{lightgray}{gray}{0.9}

\usepackage{multirow} 
\usepackage{subcaption}
\usepackage[pagebackref=true,breaklinks=true,letterpaper=true,colorlinks,bookmarks=false]{hyperref}
\usepackage[font=small]{caption}
\hypersetup{
urlcolor=magenta
}
\newcommand*\samethanks[1][\value{footnote}]{\footnotemark[#1]}

\usepackage{floatrow}
\newfloatcommand{capbtabbox}{table}[][\FBwidth]

\usepackage[table]{xcolor}
\definecolor{lightgray}{gray}{0.9}
\usepackage{multirow} 

\newcolumntype{Y}{>{\centering\arraybackslash}X}

\usepackage{color}
\begin{document}
\pagestyle{headings}
\mainmatter

\def\ECCV18SubNumber{122}  %

\titlerunning{Fighting Fake News: Image Splice Detection \\ via Learned Self-Consistency}

\authorrunning{Huh et al.}

\institute{UC Berkeley$^{1}$ \qquad \qquad Carnegie Mellon University$^{2}$}

\title{Fighting Fake News: Image Splice Detection \\ via Learned Self-Consistency}

\author{Minyoung Huh\thanks{Indicates equal contribution.}$^{1,2}$ \quad Andrew Liu\samethanks$^{1}$ \quad Andrew Owens$^{1}$ \quad Alexei A. Efros$^{1}$\\
}
\maketitle
\raggedbottom
\blfootnote{Code and additional results can be found on our~%
\href{https://minyoungg.github.io/selfconsistency/}{website}.
}

{\setlength\intextsep{0pt}
\begin{figure}
    
    \centering
    \includegraphics[width=1.0\linewidth]{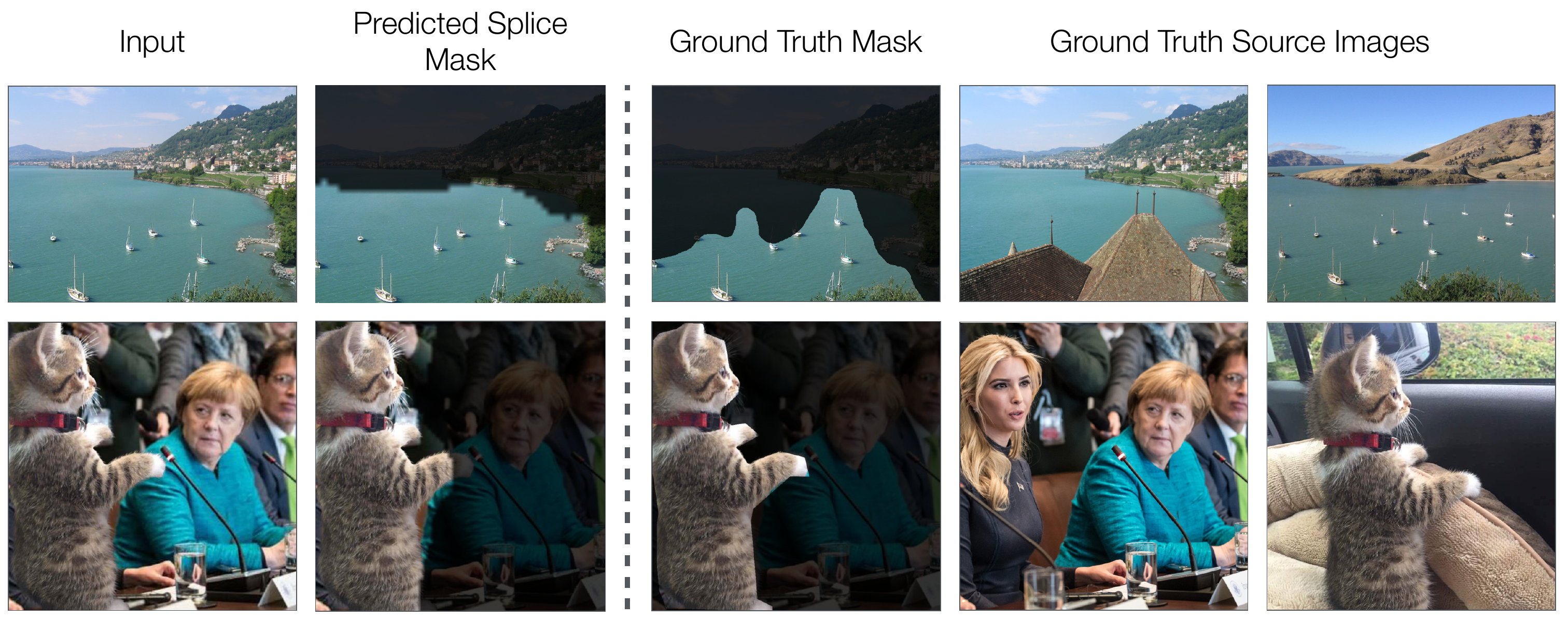}
    \vspace{-.2in}
    \caption{\small
    Our algorithm learns to detect and localize image manipulations (splices), despite being trained only on unmanipulated images. The two input images above might look plausible, but our model correctly determined that they have been manipulated because they lack self-consistency: the visual information within the predicted splice region was found to be inconsistent with the rest of the image.
    {\sc Image credits:} automatically created splice from Hays and Efros~\cite{hays2007scene} (top), manual splice from {\em Reddit} user {\em /u/Name-Albert\_Einstein} (bottom).} 
    \label{fig:teaser}
\end{figure}

 \vspace{-1.5mm}
\begin{abstract}

    Advances in photo editing and manipulation tools have made it significantly easier to create fake imagery. Learning to
    detect such manipulations, however, remains a challenging problem due to the lack of sufficient amounts of manipulated training data. 
    In this paper, we propose a learning algorithm for detecting visual image manipulations that is trained only using a large dataset of real photographs.  The algorithm uses the automatically recorded photo EXIF metadata as supervisory signal for training a model to determine whether an image is {\em self-consistent} --- that is, whether its content could have been produced by a single imaging pipeline.
    We apply this self-consistency model to the task of detecting and localizing image splices. The proposed method obtains state-of-the-art performance on several image forensics benchmarks, despite never seeing any manipulated images at training.  That said, it is merely a step in the long quest for a truly general purpose visual forensics tool.

    \vspace{-2mm}
\end{abstract}
}

\section{Introduction}

Malicious image manipulation, long the domain of dictators~\cite{king1997} and spy agencies, has now become accessible to legions of common Internet trolls and Facebook con-men~\cite{farid2016photo}.  
With only rudimentary editing skills,
it is now possible to create realistic image
composites~\cite{zhu2015composite,tsai2017deep}, fill in large image regions~\cite{hays2007scene,barnes2009patchmatch,pathak2016context}, generate plausible video from speech~\cite{suwajanakorn2017synthesizing,chung2017you}, etc. 
One might have hoped that these new methods for creating synthetic visual content would be met with commensurately powerful techniques for detecting fakes, but this has not been the case so far.

\begin{figure*}[t!]
    \centering
    \includegraphics[width=\linewidth]{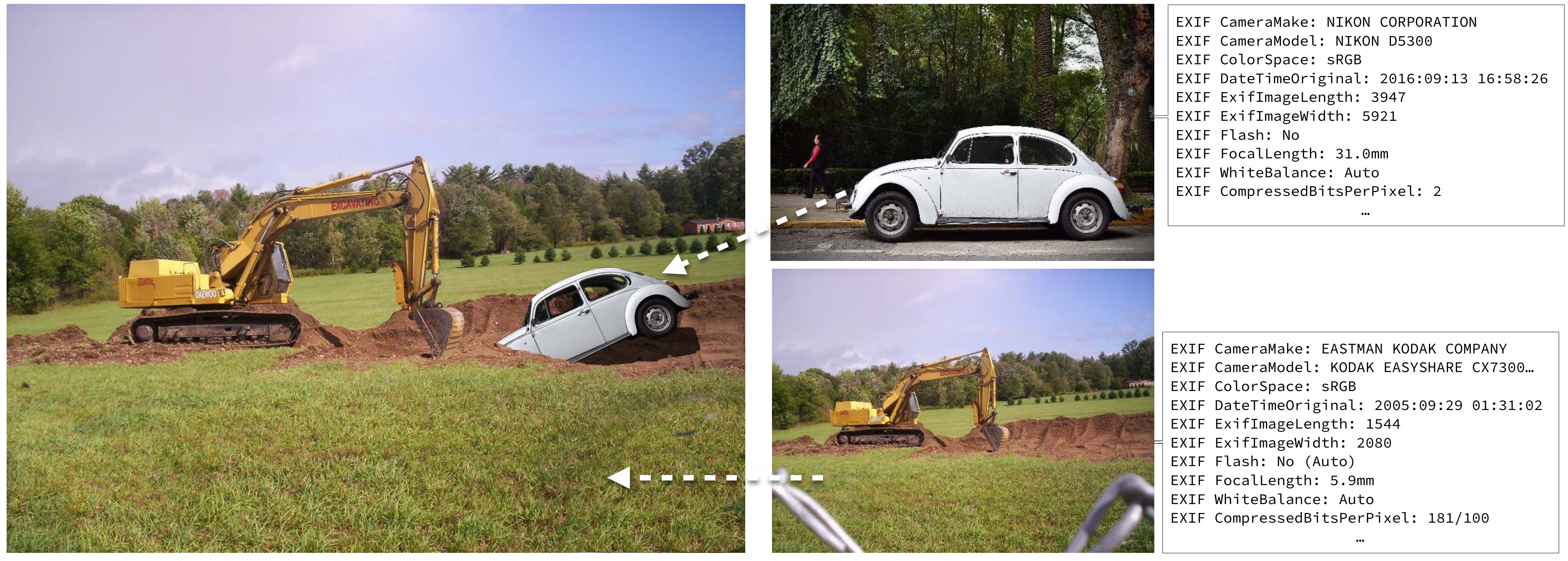}
    \vspace{-.25in}
    \caption{\small \textbf{Anatomy of a splice:} One of the most common ways of creative fake images is splicing together content from two different real source images.  %
      The insight explored in this paper is that patches from a spliced image are typically produced by different imaging pipelines, as indicated by the EXIF meta-data of the two source images.  The problem is that in practice, we never have access to these source images at test time.\protect \footnotemark}
    \label{fig:anatomy}
\end{figure*}
\footnotetext{Photo credits: NIMBLE dataset \cite{nimble2017} and \textit{Flickr} user James Stave.}
 
One problem is that standard supervised learning approaches, which have been very successful for many types of detection problems, are not well-suited for image forensics. This is because the space of manipulated images is so vast and diverse, that it is rather unlikely we will ever have enough manipulated training data for a supervised method to fully succeed.  Indeed, detecting visual manipulation can be thought of as an anomaly detection problem --- we want to flag anything that is ``out of the ordinary,'' even though we might not have a good model of what that might be.  In other words, we would like a method that does not require any manipulated training data at all, but can work in an unsupervised/self-supervised regime. 

In this work, we turn to a vast and previously underutilized source of data, image EXIF metadata. EXIF tags are camera specifications that are digitally engraved into an image file at the moment of capture and are ubiquitously available. Consider the photo shown in \fig{fig:anatomy}. While at first glance it might seem authentic, we see on closer inspection that a car has been inserted into the scene.  The content for this spliced region came from a different photo, shown on the right. Such a manipulation is called an {\em image splice}, and it is one of the most common ways of creating visual fakes.  
If we had access to the two source photographs, we would see from their EXIF metadata that there are a number of differences in the imaging pipelines: one photo was taken with an {\em Nikon} camera, the other with a {\em Kodak} camera; they were shot using different focal lengths, and saved with different JPEG quality settings, etc.  Our insight is that one might be able to detect spliced images because they are composed of regions that were captured with different imaging pipelines.
Of course, in forensics applications, we do not have access to the original source images nor, in general, the fraudulent photo's metadata.  

Instead, in this paper, we propose to use the EXIF metadata as a {\em supervisory signal} for training a classification model to determine whether an image is {\em self-consistent} -- that is, whether different parts of the same image could have been produced by a single imaging pipeline.  The model is self-supervised in that only real photographs and their EXIF meta-data are used for training.  A consistency classifier is learned for each EXIF tag separately using pairs of photographs, and the resulting classifiers are combined together to estimate self-consistency of pairs of patches in a novel input image.  We validate our approach using several datasets and show that the model performs better than the state-of-the-art --- despite never having seen annotated splices or using handcrafted detection cues.

The main contributions of this paper are: 1) posing image forensics as a problem of detecting violations in learned self-consistency (a kind of anomaly detection),
2) proposing photographic metadata as a free and plentiful supervisory signal for learning self-consistency,
3) applying our self-consistency model to detecting and localizing splices. We also introduce a new dataset of image splices obtained from the internet, and experimentally evaluate which photographic metadata is predictable from images.

\section{Related work}

Over the years, researchers have proposed a variety of visual forensics methods for identifying various manipulations~\cite{farid2016photo}. The earliest and most thoroughly studied approach is to use domain knowledge to isolate physical cues within an image. Drawing upon techniques from signal processing, previous methods focused on cues such as misaligned JPEG blocks~\cite{liu2011jpeg}, 
compression quantization artifacts~\cite{luo2010jpeg},
resampling artifacts~\cite{huang2010detecting}, color filtering array discrepancies~\cite{popescu2005exposing}, and camera-hardware ``fingerprints''~\cite{swaminathan2008fingerprint}. We take particular inspiration from recent work by Agarwal and Farid~\cite{agarwal2017dimples}, which exploits a
seemingly insignificant difference between imaging pipelines to detect spliced image regions --- namely, the way that different cameras truncate numbers during JPEG quantization. While these domain-specific approaches have proven to be useful due to their easy interpretability, we believe that the use of machine learning will open the door to discovering many more useful cues while also producing more adaptable algorithms.

Indeed, recent work has moved away from using {\em a priori} knowledge and toward applying end-to-end learning methods for solving specific forensics tasks using labeled training data. For example, Salloum \etal \cite{salloumfcn} propose learning to detect splices by training a fully convolutional network on labeled training data. These learning methods have also been applied to the problem of detecting specific tampering cues, such as double-JPEG compression~\cite{dbljpegbarni2017,amerini2017localization} and contrast enhancement~\cite{wen2017contrast}. The most closely related of these methods to ours is perhaps Bondi \etal~\cite{bondi2017camera,bondi2017tampering}. This work recognizes camera models from image patches, and proposes to use inconsistencies in camera predictions to detect tampering. Another common forensics strategy is to train models on a small class of automatically simulated manipulations, like face-swapping \cite{Davis_2017_CVPR_Workshops} or splicing with COCO segmentation masks \cite{Zhou_2018_CVPR}. In addition, \cite{Davis_2017_CVPR_Workshops} propose identifying face swaps by measuring image inconsistencies introduced from splicing and blurring. In concurrent work, Mayer~\cite{drexel2018} proposed using a Siamese network to predict whether pairs of image patches have the same camera model --- a special case of our metadata consistency model (they also propose using this model for splice detection; while promising, these results are very preliminary). There has also been work that estimates whether a photo's semantic content (e.g., weather) matches its metadata~\cite{chen2017_metadata_tampering}.

\begin{figure*}[t]
    \centering
    \includegraphics[width=1.0\linewidth]{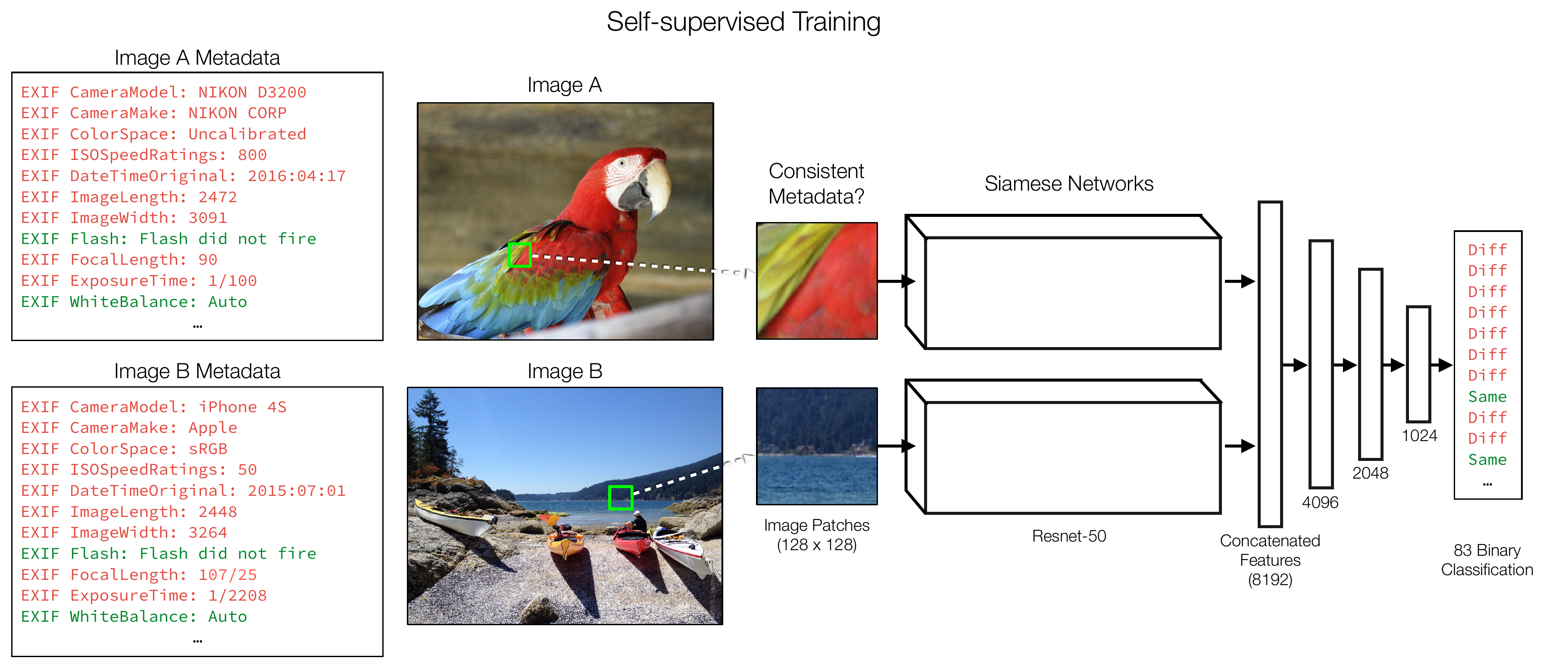}
    \caption{{\bf Self-supervised training:} Our model takes two random patches from different images and predicts whether they have consistent meta-data. Each attribute is used as a consistency metric during training and testing.}\vspace{-2mm}
    \label{fig:diagram}
\end{figure*} 
In our work, we seek to further reduce the amount of information we provide to the algorithm by having it learn to detect manipulations without ground-truth annotations. For this, we take inspiration from recent works in self-supervision~\cite{de1994learning,doersch2015unsupervised,jayaraman2015learning,agrawal2015learning,owens2016ambient,zhang2016splitbrain} which train models by solving tasks solely defined using unlabeled data. Of these, the most closely related approach is that of Doersch \etal~\cite{doersch2015unsupervised}, in which they trained a model to predict the relative position of pairs of patches within an image.  Surprisingly, the authors found that their method learned to utilize very subtle artifacts like chromatic lens aberration as a shortcut for learning the task.  While imaging noise was a nuisance in their work, it is a useful signal for us --- our self-supervised algorithm is designed to learn about properties of the imaging pipeline while ignoring semantics. Our technical approach is also similar to~\cite{isola_same_im}, which trains a segmentation model using self-supervision to predict whether pairs of patches co-occur in space or time. 

Individual image metadata tags, such as focal length, GPS, hashtags, etc. have long been employed in computer vision as free supervisory signal.
A particularly creative use of EXIF metadata was demonstrated by Kuthirummal \etal~\cite{kuthirummal2008priors}, who used the {\tt CameraModel} tag of a very large image collection to compute per-camera priors such as their non-linear response functions.%
Our work is also related to the anomaly detection problem.  Unlike traditional visual anomaly detection work, which is largely concerned with
detecting unusual semantic events like the presence of rare objects and actions~\cite{hoai2014max,mahadevan2010anomaly}, our work needs to find anomalies in photos
whose content is designed to be plausible enough to fool humans. Therefore the anomalous cues we search for should be imperceptible to humans and invariant to the semantics of the scene. 

\section{Learning Photographic Self-consistency}

Our model works by predicting whether a pair of image patches are consistent with each other. Given two patches, $\mathcal{P}_i$ and $\mathcal{P}_j$, we estimate the probabilities $x_1, x_2, ..., x_n$ that
they share the same value for each of $n$ metadata attributes. We then estimate the patches' overall
consistency, $c_{ij}$, by combining our $n$ observations of metadata consistency. 
At evaluation time, our model takes a potentially manipulated test image and measures the consistency between many different pairs of patches. A low consistency score indicates that the patches were likely produced by two distinct imaging systems, suggesting that they originate from different images.  Although the consistency score for any single pair of patches will be noisy, aggregating many observations provides a reasonably stable estimate of overall image self-consistency.

\subsection{Predicting EXIF Attribute Consistency} 
\label{sec:predexif}

We use a Siamese network to predict the probability that a pair of $128\times128$ image patches shares the same value for each EXIF metadata attribute. We train this network with image patches randomly sampled from $400,000$ \textit{Flickr} photos, making predictions on all EXIF attributes that appear in more than $50,000$ photos ($n = 80$, the full list of attributes can be found in supplementary files). For a given EXIF attribute, we discard EXIF values that occur less than $100$ times. The Siamese network uses shared ResNet-50 \cite{he2016deep} sub-networks which each produce 4096-dim. feature vectors. These vectors are concatenated and passed through four-layer MLP with 4096, 2048, 1024 units, followed by the final output layer. The network predicts the probability that the images share the same value for each of the $n$ metadata attributes.  

We found that training with random sampling is challenging because: 1) there are some rare EXIF values that are very difficult to learn, and 2) randomly selected pairs of images are unlikely to have consistent EXIF values by chance. Therefore, we introduce two types of re-balancing: unary and pairwise.  
For unary re-balancing, we oversample rare EXIF attribute values (e.g. rare camera models). When constructing a mini-batch, we first choose an EXIF attribute and uniformly sample an EXIF value from all possible values of this attribute. 
For pairwise re-balancing, we make sure that pairs of training images within a mini-batch are selected such that for a given EXIF attribute, half the batch share that value and half do not. 

\comm{
    1. do we need to have a separate analysis paragraph, lets just get rid of it.
   although we train an all available exif tags .. in hindsight we expect certain tags to be useful (e.g ?) and certain tags to be not (e.g ?). Hence, we evaluated our EXIF-consistency model on a held-out set without class balancing. Surprisingly, UserComment is the most predictable exif tag. With further inspection we see that ... Talk about lensmake now.
}
\xpar{Analysis} Although we train on all common EXIF attributes, we expect the model to excel at distinguishing ones that directly correlate to properties of the imaging pipeline such as {\tt LensMake}~\cite{doersch2015unsupervised,bondi2017camera}. In contrast, arbitrary attributes such as the exact date an image was taken ({\tt DateTimeOriginal}) leave no informative cues in an image. In order to identify predictive metadata, we evaluated our EXIF-consistency model on a dataset of 50K held-out photos and report the individual EXIF attribute accuracy in~\fig{fig:exif-acc} (chance is $50\%$ due to rebalancing). %
\begin{figure}[t!]
\begin{floatrow}
\ffigbox{
    \includegraphics[width=1.0\linewidth]{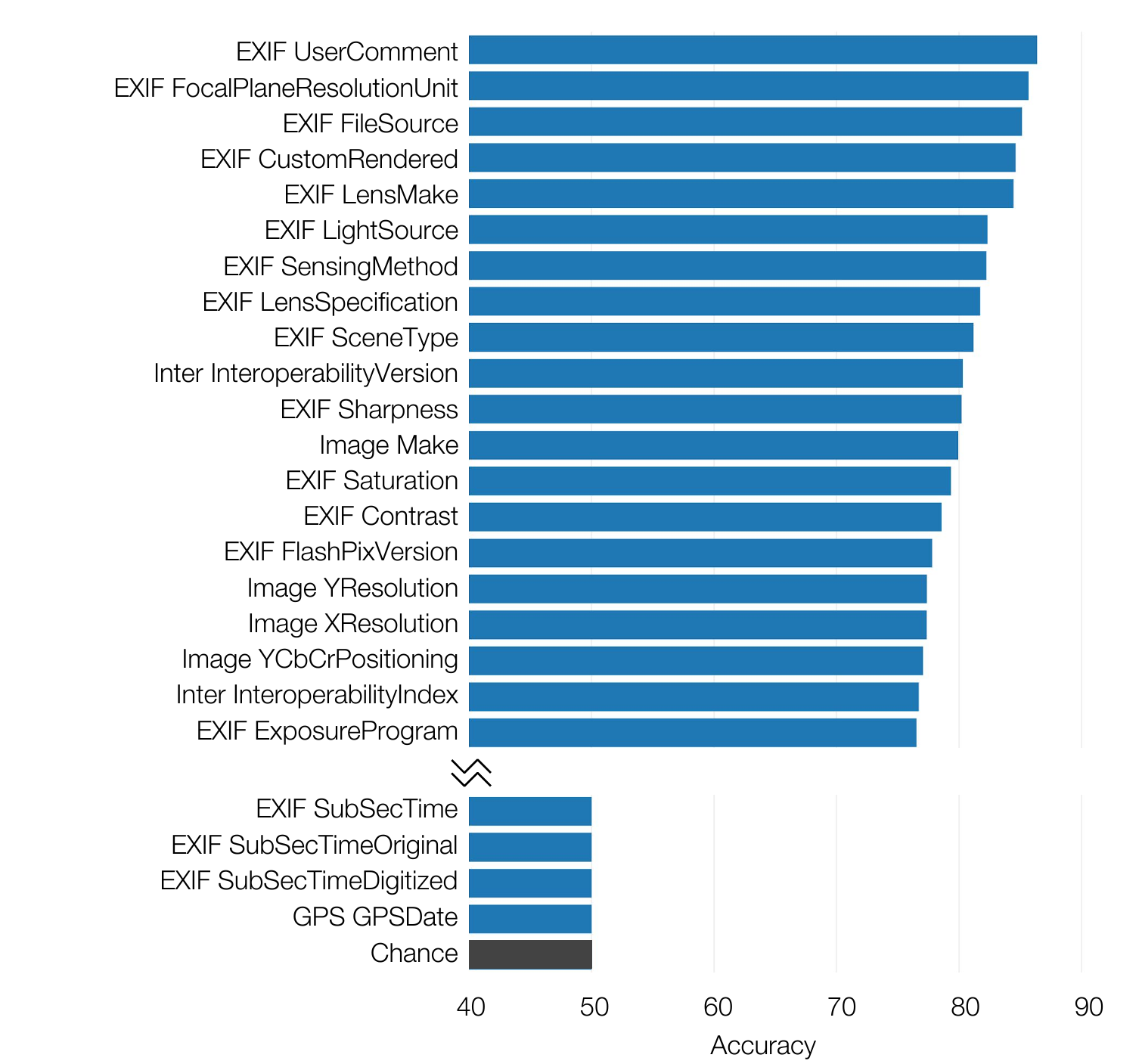}
}{
  \caption{\small \textbf{EXIF Accuracy:} How predictable are EXIF attributes? For each attribute, we compute pairwise-consistency accuracy on \textit{Flickr} images using our self-consistency model.\normalsize}
  \label{fig:exif-acc}
 }
\ffigbox{
    \includegraphics[width=0.9\linewidth]{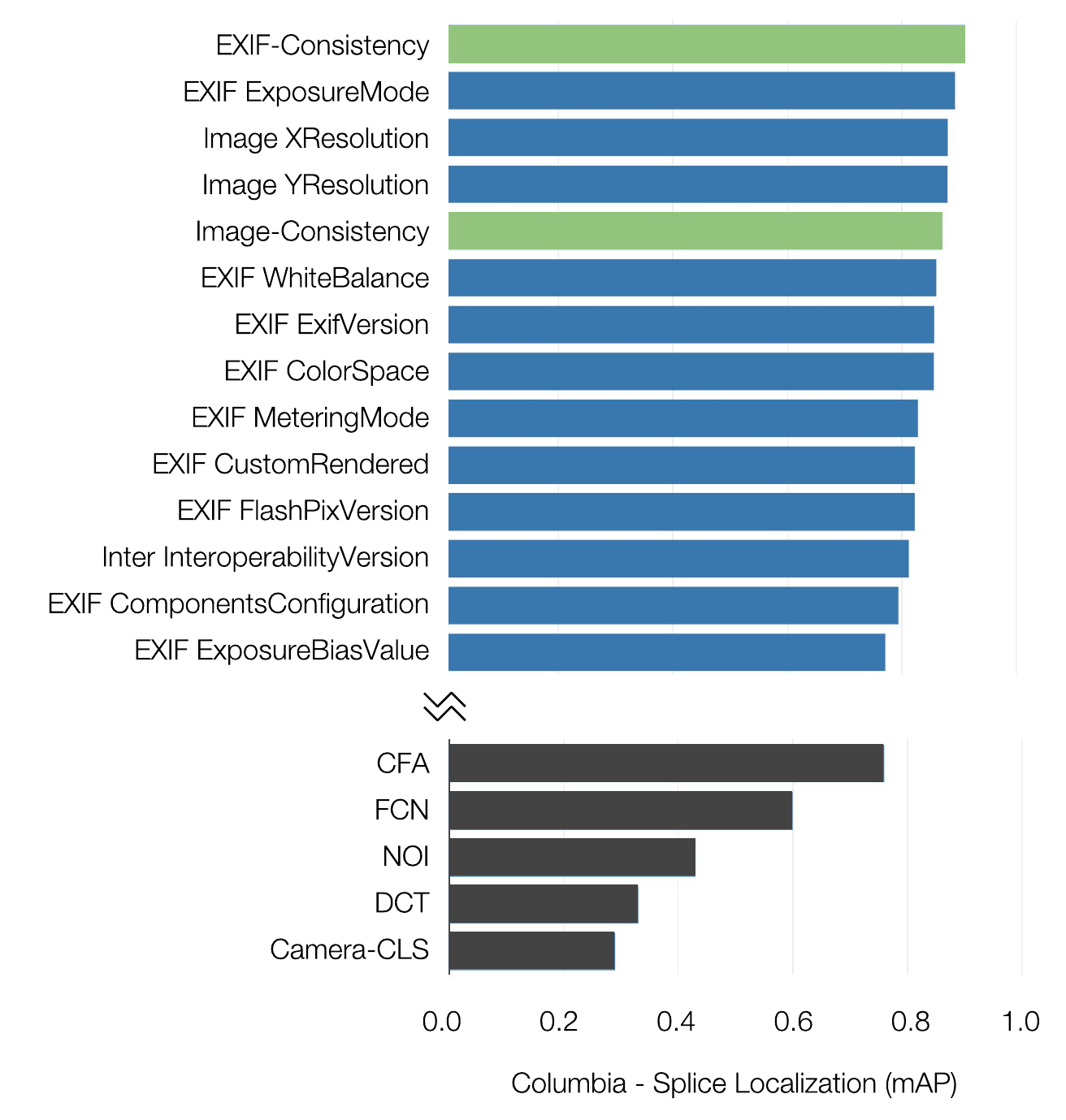}%
}{%
    \caption{\small \textbf{EXIF Splice Localization:} How useful are EXIF attributes for localizing splices? We compute individual localization scores on the \textit{Columbia} dataset. \normalsize}%
    \label{fig:exif-loc}
 }
\end{floatrow}
\end{figure}

\comm{
    
    \begin{figure}[t!]
        \begin{subfigure}[t]{0.49\linewidth}
            \includegraphics[width=0.9\linewidth]{./images/individual_acc.pdf}
            \label{fig:exif-acc}
            \subcaption{Image patch accuracy on predicting inconsistencies from the NIMBLE dataset.}
        \end{subfigure}
        \begin{subfigure}[t]{0.49\linewidth}
            \includegraphics[width=1.0\linewidth]{./images/individual_columbia.pdf}
            \subcaption{Quality of splice detection per EXIF tag on \textit{Columbia} dataset.}
            \label{fig:exif-map}
        \end{subfigure}
        \label{fig:bar}
    \end{figure}
}

 Our model obtains high accuracy when predicting the consistency of attributes closely associated with the image formation process such as {\tt LensMake}, which contains values such as {\em Apple} and {\em FUJIFILM}. But more surprisingly, we found that the most predictable attribute is {\tt UserComment}. Upon further inspection, we found that {\tt UserComment} is a generic field that can be populated with arbitrary data, and that its most frequent values were either binary strings embedded by camera manufacturers or logs left by image processing software. For example, one of its common values, {\em Processed with VSCOcam}, is added by a popular photo-filtering application. Please see the supplementary material for a full list of EXIF attributes and their definitions.

\comm{
    Our model obtains relatively high accuracy when predicting the consistency of attributes that are closely associated with the image formation process and camera model, such as {\tt LensMake} and {\tt LensSpecification}. For example, the two most common values of {\tt LensMake} are {\em Apple} and {\em FUJIFILM}. As expected, attributes that have little correlation with the image content, such as {\tt GPSDate}, were predictable only at chance level.
    Interestingly, the attribute whose consistency was most predictable, {\tt UserComment}, is a generic field that the user can fill in with arbitrary data. We found that after our label pruning process, however, the most frequent values were either binary strings embedded by the camera manufacturers, or messages left by popular image processing software. For example, one common value was {\em Processed with VSCOcam}, a popular photo-filtering phone application. We provide a complete list of tags and their common values in \fig{XXXXX}.
}

\subsection{Post-processing Consistency}

Many image manipulations are performed with the intent of making the resulting image look plausible to the human eye: spliced regions are resized, edge artifacts are smoothed, and the resulting image is re-JPEGed. If our network could predict whether two patches are post-processed differently, then this would be compelling evidence for photographic inconsistency.
To model post-processing consistency, we add three augmentation operations during training: re-JPEGing, Gaussian blur, and image resizing. Half of the time, we apply the same operations to both patches; the other half of the time, we apply different operations. The parameters of each operation are randomly chosen from an evenly discretized set of numbers.  We introduce three additional classification tasks (one per augmentation type) that are used to train the model to predict whether a pair of patches received the same parameterized augmentation. This increases the number of binary attributes we predict from 80 to 83. Since the order of the post-processing operations matters, we apply them in a random order each time. 
We note that this form of inconsistency is orthogonal to EXIF consistency. For example, in the (unlikely) event that a spliced region had exactly the same metadata as the image it was inserted into, the splice could still be detected by observing differences in post-processing.

\subsection{Combining Consistency Predictions}

Once we have predicted the consistency of a pair of patches for each of our EXIF (plus post-processing) attributes, we would like to estimate the pairs' {\em overall} consistency $c_{ij}$. If we were solving a supervised task, then a natural choice would be to use spliced regions as supervision to predict, from the $n$ EXIF-consistency predictions, the probability that the two patches belong to different regions. Unfortunately, we do not have spliced images to train on.  Instead, we use a self-supervised proxy task: we train a simple classifier to predict, from the EXIF consistency predictions, whether the patches come from the same image.

More specifically, consider the 83-dimensional vector $\vec{x}$ of EXIF consistency predictions for a pair of patches $i$ and $j$. We estimate the overall consistency between the patches as $c_{ij} = p_\theta(y \mid \vec{x})$ where $p_\theta$ is a two-layer MLP with 512 hidden units. The network is trained to predict whether $i$ and $j$ come from the same training image (\ie $y = 1$ if they're the same; $y = 0$ if they're different). This has the effect of calibrating the different EXIF predictions while modeling correlations between them. 

\begin{figure*}[t!]
    \centering
    \includegraphics[width=0.9\linewidth]{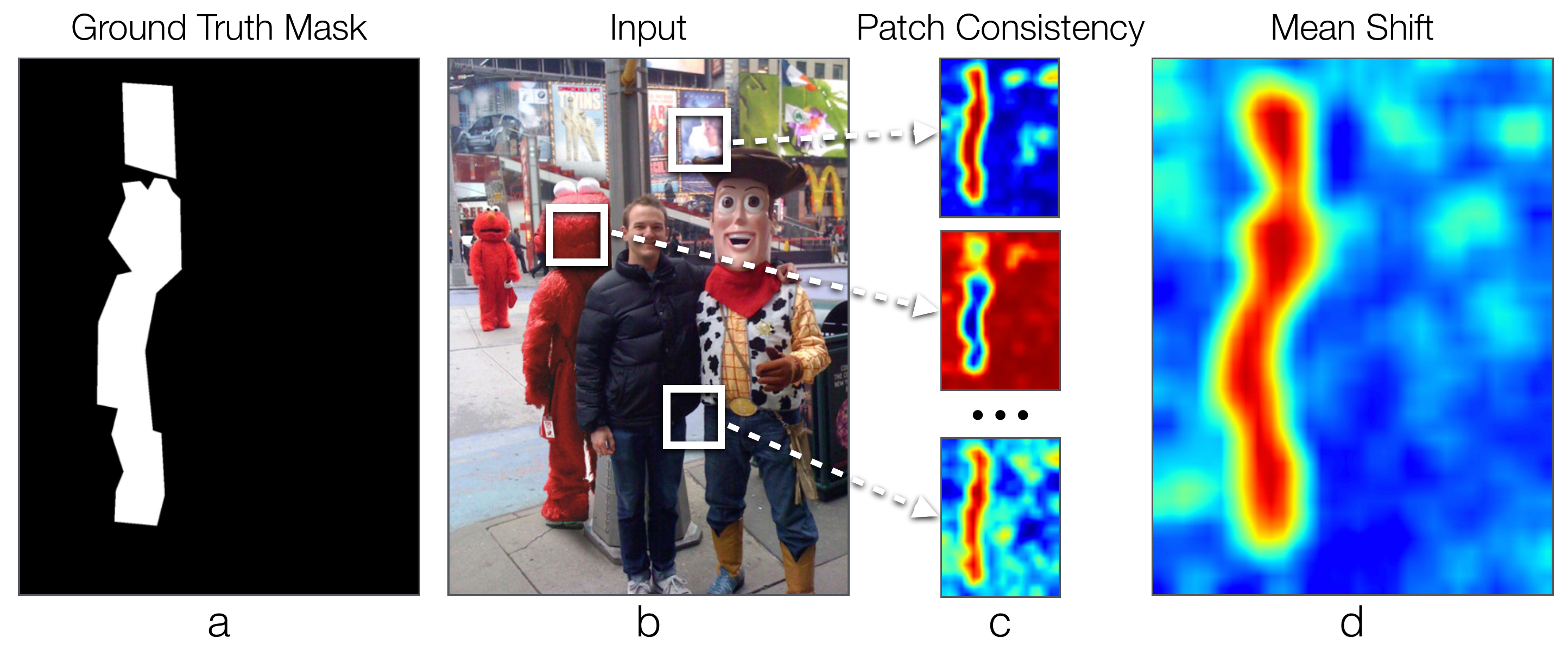}
    \caption{{\bf Test Time:} Our model samples patches in a grid from an input image (b) and estimates consistency for every pair of patches. (c) For a given patch, we get a consistency map by comparing it to all other patches in the image. (d) We use Mean Shift to aggregate the consistency maps into a final prediction.}\vspace{-2mm}
    \label{fig:test-time}
\end{figure*}

 \subsection{Directly Predicting Image Consistency}
\label{sec:imageconsistency}
An alternative to using EXIF metadata as a proxy for determining consistency between two image patches is to 
directly predict whether the two patches come from the same image or not.  Such a model could be easily trained with pairs of patches randomly sampled from the same or different images. In principle, such a model should work at least as well as the EXIF one, and perhaps better, since it could pick up on differences between images not captured by any of the EXIF tags.  In practice, however, such a model would need to be trained on vast amounts of data, because most random patches coming from different images will be easy to detect with trivial cues.  
For example, the network might simply learn to compare patch color histograms, which is a surprisingly powerful cue for same/different image classification task~\cite{lalonde2007using,isola_same_im}. 
To evaluate the performance of this model in practice, we trained a Siamese network, similar in structure to the EXIF-consistency model~(Section~\ref{sec:predexif}), to solve the task of same-or-different image consistency (see {\em Image-Consistency} in the Results section).

\subsection{From Patch Consistency to Image Self-Consistency}

So far we have introduced models that can measure some form of consistency between pairs of patches. In order to transform this into something usable for detecting splices, we need to aggregate these pairwise consistency probabilities into a global self-consistency score for the entire image.

Given an image, we sample patches in a grid, using a stride such that the number of patches sampled along the longest image dimension is 25. This results in at most 625 patches (for the common 4:3 aspect ratio, we sample $25 \times 18 = 450$ patches). For a given patch, we can visualize a response map corresponding to its consistency with every other patch in the image. To increase the spatial resolution of each response map, we average the predictions of overlapping patches.
If there is a splice, then the majority of patches from the untampered portion of the image will ideally have low consistency with patches from the tampered region
(Figure~\ref{fig:test-time}c). 

To produce a single response map for an input image, we want to find the most consistent mode among all patch response maps. 
We do this mode-seeking using Mean Shift~\cite{mshift}.
The resulting response map naturally segments the image into consistent and inconsistent regions (Figure~\ref{fig:test-time}d). We call the merged response map a {\em consistency map}. We can also qualitatively visualize the tampered image region by clustering the affinity matrix, \eg with Normalized Cuts~\cite{ncuts2000}.

\begin{figure*}[t!]
    \centering
    \includegraphics[width=1.0\linewidth]{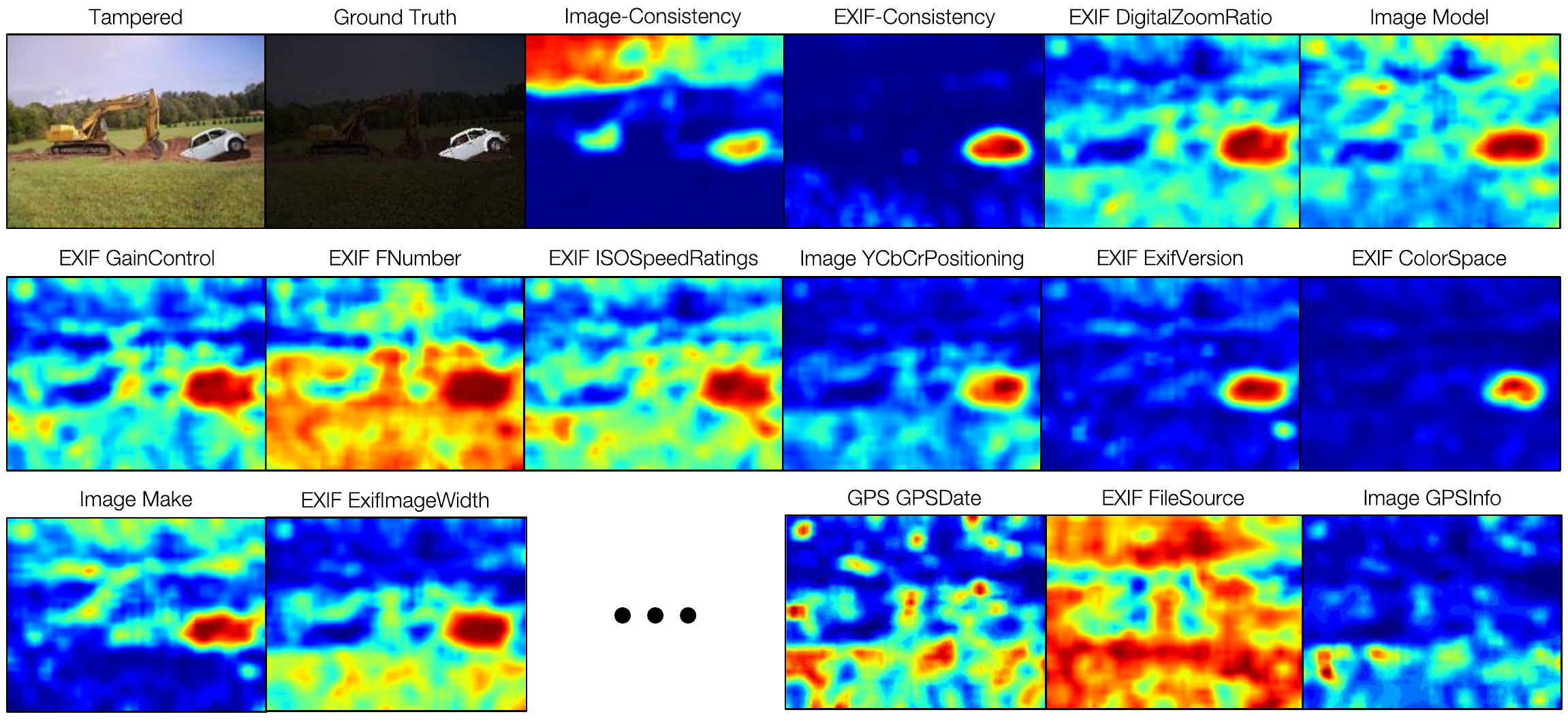}
    \caption{\textbf{Consistency map from different EXIF tags:} We compute consistency maps for each metadata attribute independently (response maps sorted by localization accuracy).  The merged consistency map accurately localizes the spliced car. }\vspace{-2mm}
    \label{fig:individual}
\end{figure*}
 To help understand how different EXIF attributes vary in their consistency predictions, we created response maps for each tag for an example image (\fig{fig:individual}). While the individual tags provide a noisy consistency signal, the merged response map accurately localizes the spliced region.

\vspace{-2mm}
\section{Results}

We evaluate our models on two closely related tasks: splice detection and splice localization.  In the former, our goal is to classify images as being spliced {\em vs.} authentic. In the latter, the goal is to localize the spliced regions within an image. %

\subsection{Benchmarks}

We evaluate our method on five different datasets. This includes three existing datasets: the widely used \textit{Columbia} dataset ~\cite{columbiadata}, which consists of $180$ relatively simple splices, and two more challenging datasets, \textit{Carvalho} \etal~\cite{carvalhodata} ($94$ images) and \textit{Realistic Tampering}~\cite{realistictampering} ($220$ images), which combine splicing with post-processing operations. The latter also includes other tampering operations, such as copy-move. %

One potential shortcoming of these existing datasets is that they were created by a small number of artists and may not be representative of the variety of forgeries encountered online.  To address this issue, we introduce a new \textit{In-the-Wild} forensics dataset that consists of $201$ images scraped from  {\sc The Onion}, a parody news website (i.e. fake news), and {\sc Reddit Photoshop Battles}, an online community of users who create and share manipulated images (which has been used in other recent forensics work \cite{moreira2018image}). Since ground truth labels are not available for internet splices, we annotated the images by hand to obtain approximate ground truth (using the unmodified source images as reference when they were available). 

Finally, we also want to evaluate our method on automatically-generated splices. For this, we used the scene completion data from Hays and Efros~\cite{hays2007scene}, which comes with inpainting results, masks, and source images for a total of $55$ images. We note that the ground-truth masks are only approximate, since the scene completion algorithm may alter a small region of pixels outside the mask in order to produce seamless splices.

\setlength{\tabcolsep}{4pt}
\begin{SCtable}[\sidecaptionrelwidth][t]
\begin{minipage}[t!]{0.55\linewidth}
\small
\centering 
\scalebox{0.7}{
\begin{tabular}{l c c c} 
    \toprule
    Dataset  & 
    \textbf{Columbia \cite{columbiadata}} & 
    \textbf{Carvalho \cite{carvalhodata}} &  
    \textbf{RT \cite{realistictampering}} \\
    
    \midrule %
    CFA \cite{cfa}         & 0.83 & 0.64 & 0.54 \\
    DCT \cite{DCT}         & 0.58 & 0.63 & 0.52 \\
    NOI \cite{NOI}         & 0.73 & 0.66 & 0.52 \\
    \midrule %
    Supervised FCN         & 0.57 & 0.56 & \textbf{0.56} \\
    \midrule
    Camera Classification  & 0.70 & 0.73 & 0.15  \\
    X-Consistency  & 0.47 & 0.46 & 0.53  \\
    Y-Consistency  & 0.48 & 0.42 & 0.56  \\
    Image-Consistency            & 0.97 & 0.75 & 0.58 \\
    EXIF-Consistency       & \textbf{0.98} & \textbf{0.87} & 0.55 \\
    \bottomrule
\end{tabular}
}
\end{minipage}\hfill
\begin{minipage}[t!]{0.45\linewidth}
\caption{\textbf{Splice Detection:} We compare our splice detection accuracy on 3 datasets. We measure the mean average precision (mAP) of detecting whether an image has been spliced. We note that RT is a dataset that contains a variety of manipulations (not just splicing).
}
\label{table:splice_detection}
\end{minipage}
\end{SCtable}
\setlength{\tabcolsep}{4pt}

\subsection{Comparisons}
We compared our model with three methods that use image processing techniques to detect specific imaging artifacts: Color Filter Array (CFA)~\cite{cfa} detects artifacts in color pattern interpolation; JPEG DCT~\cite{DCT} detects inconsistencies over JPEG coefficients; and Noise Variance (NOI) ~\cite{NOI} detects anomalous noise patterns using wavelets. We used implementations of these algorithms provided by Zampoglou et al.~\cite{zamp16}.

Since we also wanted to compare our unsupervised method with approaches that were trained on labeled data, we report results from a learning-based method: {E-MFCN} \cite{salloumfcn}.  Given a dataset of spliced images and masks as training data, they use a supervised fully convolutional network (FCN)~\cite{fcn8shelhamer} to predict splice masks and boundaries in test images.  
To test on our new datasets, we implemented a simplified version of their model (a standard FCN trained to recognize spliced pixels) that was trained with a training split of the \textit{Columbia}, \textit{Carvalho}, and \textit{Realistic Tampering} datasets.
We split every dataset in half to construct train/test sets.

Finally, we present three variations of self-consistency models. The first, \textit{Camera-Classification}, was trained to directly predict which camera model produced a given image patch. We evaluate the output of the camera classification model by sampling image patches from a test image and assigning the most frequently predicted camera as the natural image and everything else as the spliced region. We consider an image to be untampered when every patch's predicted camera model is consistent.

The second model, \textit{XY-Consistency} learns to predict whether patches are spatially consistent -- given a pair of patches sampled in a certain order, does the network find the ordering consistent. \textit{XY-Consistency} is inspired by Doersch \etal ~\cite{doersch2015unsupervised}, where they found that a network can use chromatic aberration to predict relative location. We train this model by sampling patches from the same image, using their XY ordering as supervision. Unlike other models, the order of patches matter. During testing, we feed patches such that the model always sees the consistent ordering of test patches and self-consistency is scored using the network's prediction of consistency.

Finally,  
\textit{Image-Consistency}, is a network that directly predicts whether two patches are sampled from the same image (\sect{sec:imageconsistency}).  An image is considered likely to have been tampered if its constituent patches are predicted to have come from different images. The evaluations of these models are performed the same way as our full \textit{EXIF-Consistency} model. 

We trained our models, including the variations, using a ResNet50~\cite{he2016deep} pretrained on ImageNet~\cite{imagenet}. All networks were trained using post-processing augmentation, but only \textit{EXIF-Consistency} used post-processing consistency. We used a batch size of 128 and optimized our objective using Adam~\cite{kingmaadam} with a learning rate of $10^{-4}$. We report our results after training for 1 million iterations. The 2-layer MLP used to compute patch consistency on top of the \textit{EXIF-Consistency} model predictions was trained for $10,000$ iterations.

\vspace{-0.5mm}
\subsection{Splice Detection}

We evaluate splice detection using the three datasets that contain both untampered and manipulated images: \textit{Columbia}, \textit{Carvalho}, and \textit{Realistic Tampering}. For each algorithm, we extract the localization map and obtain an overall score by spatially averaging the responses. The images are ranked based on their overall scores, and we compute the mean average precision (mAP) for the whole dataset. %

\tbl{table:splice_detection} shows the mAP for detecting manipulated images. Our \textit{Consistency} models achieves state-of-the-art performance on \textit{Columbia} and \textit{Carvalho} and Realistic Tampering, beating supervised methods like \textit{FCN}.  %

\begin{figure*}[t!]
    \centering
    \includegraphics[width=.98\textwidth,keepaspectratio]{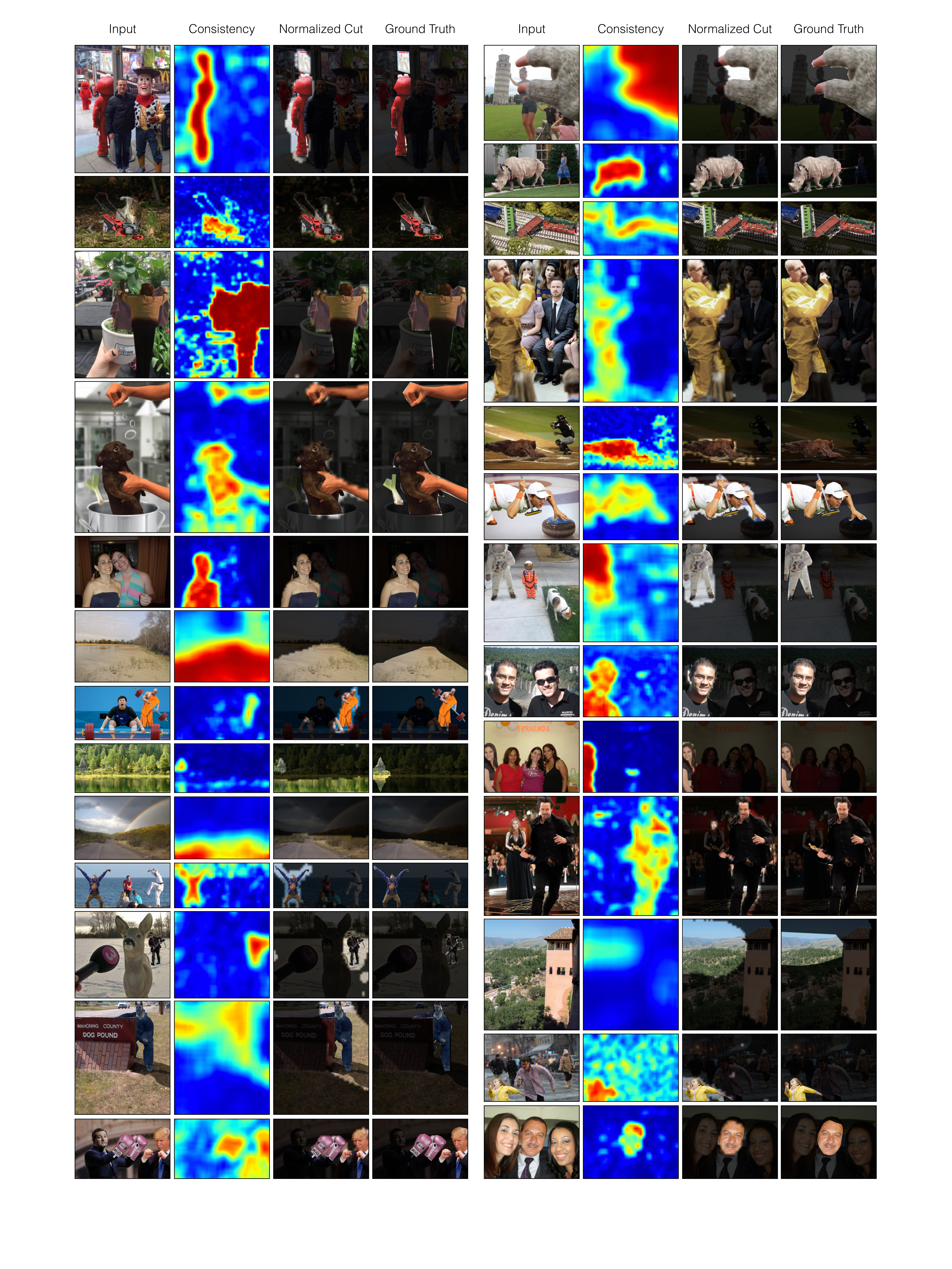}
    
    \caption{\textbf{Detecting Fakes:} \textit{EXIF-Consistency} successfully localizes manipulations across many different datasets. We show qualitative results on images from {\em Carvalho}, {\em In-the-Wild}, {\em Hays} and {\em Realistic Tampering}.}\vspace{-2mm}
    \label{fig:all_results}
\end{figure*}
\begin{figure*}[t!]
    \centering
    \includegraphics[width=1.0\linewidth]{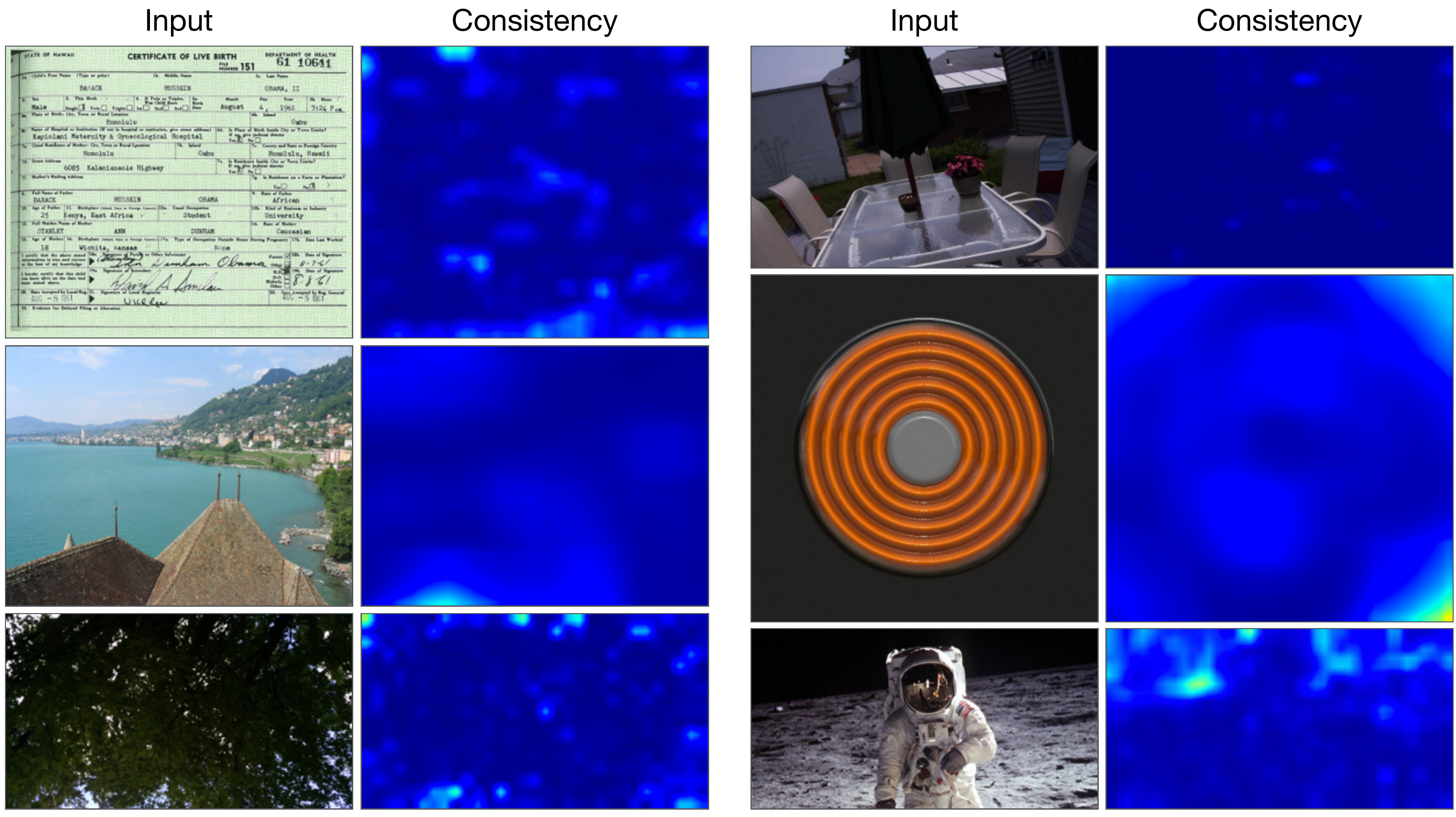}
    \caption{\textbf{Response on Untampered Images:} Our algorithm's response map contains fewer inconsistencies when given an untampered images.}\vspace{-2mm}
    \label{fig:neutral-results}
\end{figure*}
\begin{figure*}[t!]
    \centering
    \includegraphics[width=1.0\linewidth]{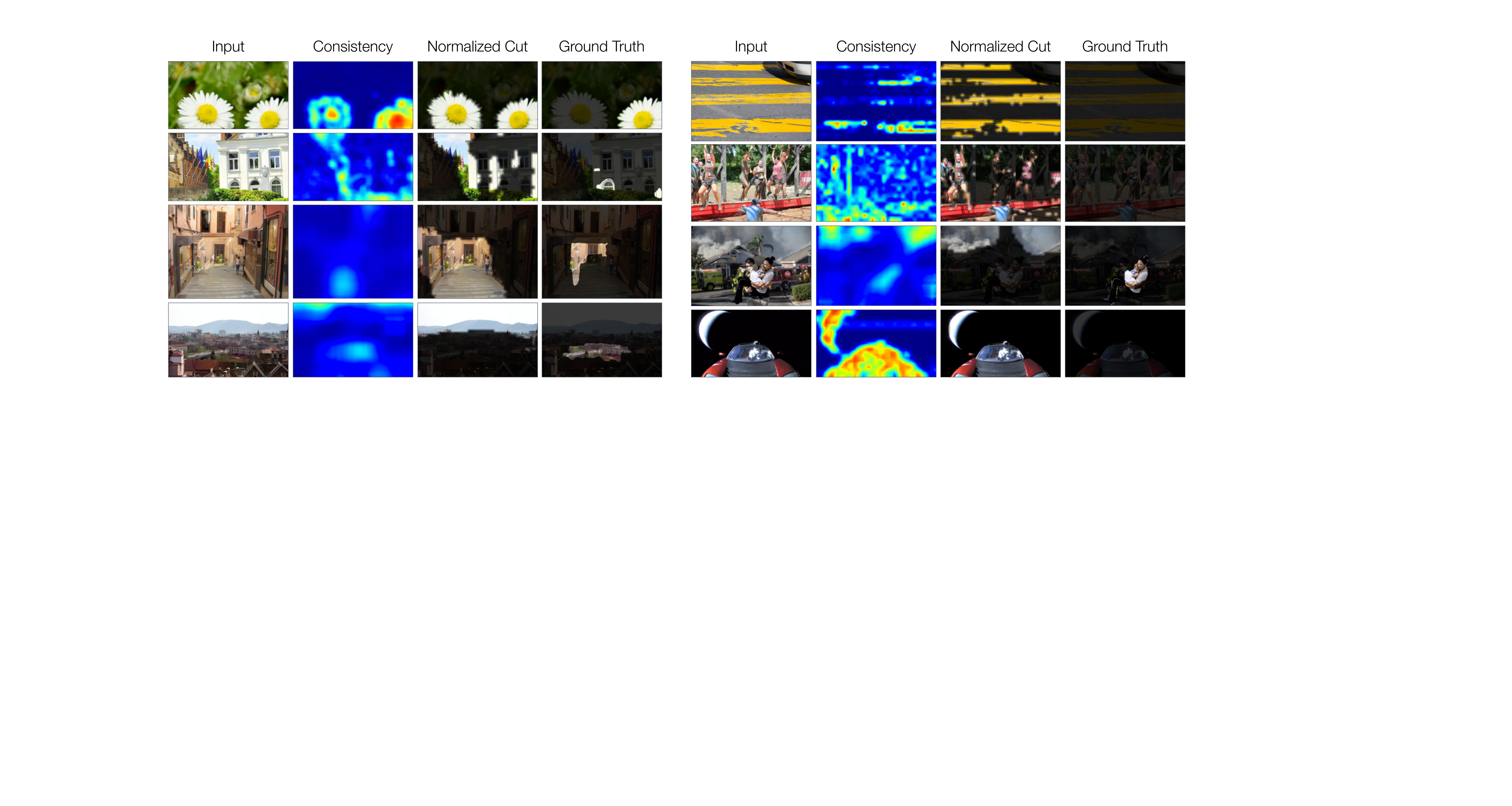}
    \caption{\textbf{Failure Cases:} We present typical failure modes of our model. As we can see with outdoor images, overexposure frequently leads to false positives in the sky. In addition some splices are too small that we cannot effectively locate them using consistency. Finally, the flower example produces a partially incorrect result when using the \textit{EXIF Consistency} model. Since the manipulation was a copy-move, the manipulation is only detectable via post-processing consistency cues (and not EXIF-consistency cues).}\vspace{-2mm}
    \label{fig:failure-results}
\end{figure*}

\setlength{\tabcolsep}{4pt}
\begin{table}[t!]
\centering
\scalebox{0.63}{
    \begin{tabular}{lccc|ccc|ccc|ccc|ccc} 
    \toprule
    Dataset & 
    \multicolumn{3}{c}{\textbf{Columbia}~\cite{columbiadata}} & 
    \multicolumn{3}{c}{\textbf{Carvalho}~\cite{carvalhodata}} &
    \multicolumn{3}{c}{\textbf{RT}~\cite{realistictampering}} &
    \multicolumn{3}{c}{\textbf{In-the-Wild}} &
    \multicolumn{3}{c}{\textbf{Hays}~\cite{hays2007scene}} \\
    \midrule
    Metric & {mAP} & {p-mAP} & {cIOU} & {mAP} & {p-mAP} & {cIOU} & {mAP} & {p-mAP} & {cIOU} & {mAP} & {p-mAP} & {cIOU} & {mAP} & {p-mAP} & {cIOU}\\ 
    \midrule  %
    CFA~\cite{cfa}           & 0.76 & 0.76 & 0.75 & 0.18 & 0.24 & 0.46 & \textbf{0.40} & \textbf{0.40} & \textbf{0.63} & 0.23 & 0.27 & 0.45 & 0.11 & 0.22 & 0.45 \\
    DCT~\cite{DCT}           & 0.33 & 0.43 & 0.41 & 0.25 &  0.32 & 0.51 & 0.11 & 0.12 & 0.50 & 0.35 &0.41& 0.51 & 0.16 & 0.21 & 0.47 \\
    NOI~\cite{NOI}           & 0.43 & 0.56 & 0.47 & 0.23 & 0.38 &0.50 & 0.12 & 0.19&  0.50 & 0.35 & 0.42 & 0.52 & 0.15 & 0.27 & 0.47\\
    \midrule %
    Supervised FCN           & 0.60 & 0.61 & 0.58 & 0.18 &  0.22 & 0.47 & 0.09 & 0.10 & 0.49 & 0.25 & 0.26& 0.46 & 0.15 & 0.17 & 0.46   \\
    \midrule %
    Camera Classification    & 0.29 & 0.65 & 0.41 & 0.11 &  0.29 &  0.44 & 0.07 & 0.10 & 0.48 & 0.20 & 0.31 & 0.44 & 0.15 & 0.31 & 0.47 \\
    X-Consistency            & 0.37 & 0.47 & 0.44 & 0.17 & 0.29 & 0.47 & 0.07 & 0.11 & 0.48 & 0.17 & 0.24 & 0.43 & 0.13 & 0.23 & 0.45 \\
    Y-Consistency             & 0.39 & 0.52 & 0.45 & 0.16 & 0.20 & 0.45 & 0.9 & 0.14 & 0.49 & 0.19 & 0.27 & 0.43 & 0.19 & 0.25 & 0.49 \\
    Image-Consistency        & 0.87 & 0.90 & 0.80 & 0.36 &  0.41 & 0.55 & 0.21 & 0.21 & 0.54 & 0.47 & \textbf{0.53} & \textbf{0.59} & 0.21 &0.37& 0.54 \\ 
    EXIF-Consistency         & \textbf{0.91}  & \textbf{0.94} & \textbf{0.85} & \textbf{0.51} & \textbf{0.52} & \textbf{0.63} & 0.20 & 0.20 & 0.54 & \textbf{0.48} & 0.49 & 0.58 & \textbf{0.48} & \textbf{0.52} & \textbf{0.65} \\

    \bottomrule
\end{tabular}
}
\caption{\textbf{Splice Localization:} We evaluate our model on 5 datasets using mean average precision (mAP, permuted-mAP) over pixels and class-balanced IOU (cIOU) selecting the optimal threshold per image.}\vspace{-1.5mm} %
\label{splice_loc}
\end{table}
\setlength{\tabcolsep}{1.4pt}

\comm{
\setlength{\tabcolsep}{4pt}
\begin{table}[t]
\begin{center}
\scalebox{0.8}{
    \begin{tabular}{lccccc} 
    \toprule
    Dataset & 
    \textbf{Columbia}~\cite{columbiadata} & 
    \textbf{Carvalho}~\cite{carvalhodata} &
    \textbf{RT}~\cite{realistictampering} &
    \textbf{In-the-Wild} &
    \textbf{Hays}~\cite{hays2007scene} \\
    \midrule  %
    CFA~\cite{cfa}           & 0.84 & 0.58 & \textbf{0.69} & 0.59 & 0.58 \\
    DCT~\cite{DCT}           & 0.58 & 0.60 & 0.53 & 0.63 & 0.52 \\
    NOI~\cite{NOI}           & 0.67 & 0.64 & 0.57 & 0.64 & 0.59 \\
    \midrule %
    Supervised FCN           & 0.80 & 0.58 & 0.58 & 0.59 & 0.57   \\
    \midrule %
    Camera Classification    & 0.59 & 0.53 & 0.51 & 0.53 & 0.55 \\
    Image-Consistency        & 0.91 & 0.67 & 0.58 & 0.69 & 0.69 \\ 
    EXIF-Consistency         & \textbf{0.97}  & \textbf{0.75} & 0.59 & \textbf{0.72} & \textbf{0.75} \\ 
    \bottomrule
\end{tabular}
}
\caption{\textbf{Splice Localization:} We evaluate our model on 5 datasets using a mean average precision-based metric (p-mAP).} %
\label{splice_loc}
\end{center}
\end{table}
\setlength{\tabcolsep}{1.4pt}
}

\comm{
    \begin{tabular}{lccc|ccc|ccc} 
    \toprule
    & \multicolumn{3}{c}{\textbf{Columbia}~\cite{columbiadata}} & 
    \multicolumn{3}{c}{\textbf{Carvalho}~\cite{carvalhodata}}&
    \multicolumn{1}{c}{\textbf{RT}~\cite{realistictampering}}&
    \multicolumn{1}{c}{\textbf{In-the-Wild}}&
    \multicolumn{1}{c}{\textbf{Hays}~\cite{hays2007scene}}\\
    \midrule
    Model                    & {MCC} & {F1} & {p-mAP} & {MCC} & {F1} & {p-mAP} & {p-mAP} & {p-mAP} & {p-mAP}\\ 
    \midrule  %
    CFA~\cite{cfa}           & 0.23 & 0.47 & 0.87 & 0.16 & 0.29 & 0.59 & \textbf{0.70} & 0.59 & 0.60 \\
    DCT~\cite{DCT}           & 0.33 & 0.52  & 0.60 & 0.19 & 0.31  & 0.60 & 0.53 & 0.63 & 0.56 \\
    NOI~\cite{NOI}           & 0.41 & 0.57  & 0.69 & 0.25 & 0.34   & 0.60 & 0.58 & 0.65 & 0.60 \\
    \midrule %
    Supervised FCN           & 0.37 & 0.57 & 0.80 & 0.07 & 0.57 & 0.58 & 0.58 & 0.59 & 0.57   \\
    \midrule %
    Camera Classification    & 0.29 & 0.51 & 0.59 & 0.15 & 0.55 & 0.53 & 0.51 & 0.54 & 0.55 \\
    Image-Consistency        & 0.55 & 0.59 & 0.91 & 0.20 & 0.76 & 0.67 & 0.58 & 0.69 & 0.69 \\ 
    EXIF-Consistency         & \textbf{0.69} & \textbf{0.71} & \textbf{0.97} & \textbf{0.37} & \textbf{0.91} & \textbf{0.75} & 0.59 & \textbf{0.72} & \textbf{0.75} \\ 
    \bottomrule
\end{tabular}
}

\comm{
    Self-Consistency(3M) & \textbf{0.68} & \textbf{0.70} & \textbf{0.97} & 0.38 & \textbf{0.90} & \textbf{0.76} & 0.59 & \textbf{0.74} & \textbf{0.74} \\ 

    \midrule
    SC(3M) + SI(700k)  & - & - & - & - & - & - & 0.594 & 0.750 & - \\ 
    Same Image(\approx 200K) & 0.45 & 0.56  & 0.87 & 0.23 & 0.72 & 0.64 & 0.58 & 0.66 & 0.66\\
        Same Image(700K)  & 0.53 & 0.58 & 0.90 & 0.23 & 0.69 & 0.67 & 0.59 & 0.71 & 0.69 \\ 
        
    Self-Consistency(700K)  & 0.54 & 0.61 & 0.89 & 0.32 & 0.76 & 0.72 & 0.58 & 0.70 & 0.66 \\

    Same Image & 0.45 & 0.56  & 0.87 & 0.23 & 0.72 & 0.64 & 0.58 & 0.66 & \textbf{0.66} \\ 
        Same Image (new) & 0.56 & 0.61 & 0.91 & - & - & - & - & 0.72 & - \\ 
            Self-Consistency  & \textbf{0.54} & \textbf{0.61}  & \textbf{0.89} & 0.32 & \textbf{0.76} & \textbf{0.72} & 0.58 & \textbf{0.70} & \textbf{0.66} \\ 
        Self-Consistency(new+dbscan)  & 0.69 & 0.72 & 0.97 & 0.38 & 0.89 & 0.77 & 0.59 & 0.76 & 0.74 \\ 
        Self-Consistency(new+ms)  & 0.68 & 0.70 & 0.97 & 0.38 & 0.90 & 0.76 & 0.59 & 0.74 & 0.74 \\ 
}

\setlength{\tabcolsep}{4pt}
\begin{SCtable}[\sidecaptionrelwidth][t!]
\begin{minipage}[t!]{0.5\linewidth}
\small
\centering 
\scalebox{0.7}{
    \begin{tabular}{lcc|cc} 
    \toprule
    Dataset &  \multicolumn{2}{c}{\textbf{Columbia}~\cite{columbiadata}} & 
    \multicolumn{2}{c}{\textbf{Carvalho}~\cite{carvalhodata}}\\
    \midrule
    Metric                     &  {MCC} & {F1} &{MCC} & {F1}\\ 
    \midrule %
     CFA \cite{cfa}            & 0.23 & 0.47 & 0.16 & 0.29 \\
     DCT \cite{DCT}            & 0.33 & 0.52 & 0.19 & 0.31 \\
     NOI \cite{NOI}            & 0.41 & 0.57 & 0.25 & 0.34 \\
    \midrule %
    E-MFCN \cite{salloumfcn}  & 0.48 & 0.61 & 0.41 & 0.48 \\
    \midrule
    Camera Classification     & 0.30 & 0.50 & 0.11 & 0.24 \\
    X-Consistency             & 0.25 & 0.54 & 0.12 & 0.30 \\
    Y-Consistency             & 0.25 & 0.54 & 0.14 & 0.28 \\
    Image-Consistency         & 0.77 & 0.85 & 0.33 & 0.43 \\
    EXIF-Consistency          & \textbf{0.80} & \textbf{0.88} &  \textbf{0.42} & \textbf{0.52}\\
    \bottomrule
\end{tabular}
}
\end{minipage}\hfill
\begin{minipage}[t!]{0.5\linewidth}
\caption{\textbf{Comparison with Salloum et al.:} We compare against numbers reported by~\cite{salloumfcn} for splice localization. } \label{table:splice_usc}
\end{minipage}
\end{SCtable}
\setlength{\tabcolsep}{4pt}

\comm{
\setlength{\tabcolsep}{1.4pt}

\setlength{\tabcolsep}{4pt}
\begin{table}[t]
\begin{center}
\scalebox{0.8}{
    \begin{tabular}{lcc|cc} 
    \toprule
    Dataset &  \multicolumn{2}{c}{\textbf{Columbia}~\cite{columbiadata}} & 
    \multicolumn{2}{c}{\textbf{Carvalho}~\cite{carvalhodata}}\\
    \midrule
    Metric                     &  {MCC} & {F1} &{MCC} & {F1}\\ 
    \midrule %
     CFA \cite{cfa}            & 0.66 (0.23) & 0.78 (0.47) & 0.10 (0.16) & 0.28 (0.29) \\
     DCT \cite{DCT}            & 0.12 (0.33) & 0.48 (0.52) & 0.20 (0.19) & 0.35 (0.31) \\
     NOI \cite{NOI}            & 0.43 (0.41) & 0.55 (0.57) & 0.18 (0.25) & 0.33 (0.34) \\
    \midrule %
    E-MFCN \cite{salloumfcn}  & (0.48) & (0.61) & (0.41) & (0.48)  \\
    \midrule
    Camera Classification     & 0.30 & 0.50 & 0.13 & 0.26 \\
    Image-Consistency         & 0.71 & 0.81 & 0.29 & 0.40 \\
    EXIF-Consistency          & \textbf{0.80} & \textbf{0.88} & \textbf{0.42} & \textbf{0.52} \\ 
    \bottomrule
\end{tabular}
}
\caption{\textbf{Comparison against Salloum et. al:} We compare against numbers reported by~\cite{salloumfcn} for splice localization. }
\label{table:splice_usc}
\end{center}
\end{table}
\setlength{\tabcolsep}{1.4pt}
}

\comm{
\setlength{\tabcolsep}{4pt}
\begin{table}[t]
\begin{center}
\scalebox{0.8}{
    \begin{tabular}{lcc|cc} 
    \toprule
    Dataset &  \multicolumn{2}{c}{\textbf{Columbia}~\cite{columbiadata}} & 
    \multicolumn{2}{c}{\textbf{Carvalho}~\cite{carvalhodata}}\\
    \midrule
    Metric                     &  {MCC} & {F1} &{MCC} & {F1}\\ 
    \midrule %
     CFA \cite{cfa}            & 0.23 & 0.47 & 0.16 & 0.29 \\
     DCT \cite{DCT}            & 0.33 & 0.52 & 0.19 & 0.31 \\
     NOI \cite{NOI}            & 0.41 & 0.57 & 0.25 & 0.34 \\
    \midrule %
    E-MFCN \cite{salloumfcn}  & 0.48 & 0.61 & \textbf{0.41} & 0.48  \\
    \midrule
    Camera Classification     & 0.29 & 0.51 & 0.15 & 0.55 \\
    Image-Consistency         & 0.58 & 0.62 & 0.17 & 0.71 \\
    EXIF-Consistency          & \textbf{0.70} & \textbf{0.72} & 0.32 & \textbf{0.86} \\ 
    \bottomrule
\end{tabular}
}
\caption{\textbf{Comparison against Salloum et. al:} We compare against numbers reported by~\cite{salloumfcn} for splice localization. We note that they compute their numbers after finding individual threshold that optimizes their score per image, while we threshold our probabilities at 0.5.}
\label{table:splice_usc}
\end{center}
\end{table}
\setlength{\tabcolsep}{1.4pt}
}

\comm{ %
    \setlength{\tabcolsep}{4pt}
    \begin{table}[t]
    \begin{center}
    \scalebox{0.8}{
        \begin{tabular}{lcccccc} 
        \toprule
        & \multicolumn{3}{c}{\textbf{Columbia}} & \multicolumn{3}{c}{\textbf{Carvalho}}\\
        \midrule
        Model & {MCC} & {F1} & {IOU}  & {MCC} & {F1} & {IOU}\\ 
        \midrule %
        CFA2    & 0.328 & 0.503 & - & 0.198 & 0.312 & - \\
        DCT     & 0.326 & 0.520 & - & 0.189 & 0.307 & - \\
        NOI1    & 0.411 & 0.574 & - & 0.245 & 0.343 & - \\
        \midrule %
        SFCN    & 0.420 & 0.582 & - & 0.368 & 0.441 & - \\
        MFCN    & 0.465 & 0.604 & - & 0.390 & 0.468 & - \\
        E-MFCN  & 0.479 & 0.612 & - & {0.407} & 0.480 & - \\
        \midrule
        Ours (image)     & 0.534 & 0.614 & 0.487  & 0.246 & 0.395 & 0.276 \\ 
        Ours (camera)    & \textbf{0.611} & \textbf{0.728} & \textbf{0.612}  & 0.242 & 0.334 & 0.237 \\
        Camera class     &  0.326 & 0.566 & 0.411 \\
        \bottomrule
    \end{tabular}
    }
    \end{center}
    \end{table}
    \setlength{\tabcolsep}{1.4pt}
} 
\begin{figure*}[t]
    \centering
    \includegraphics[width=1.0\linewidth]{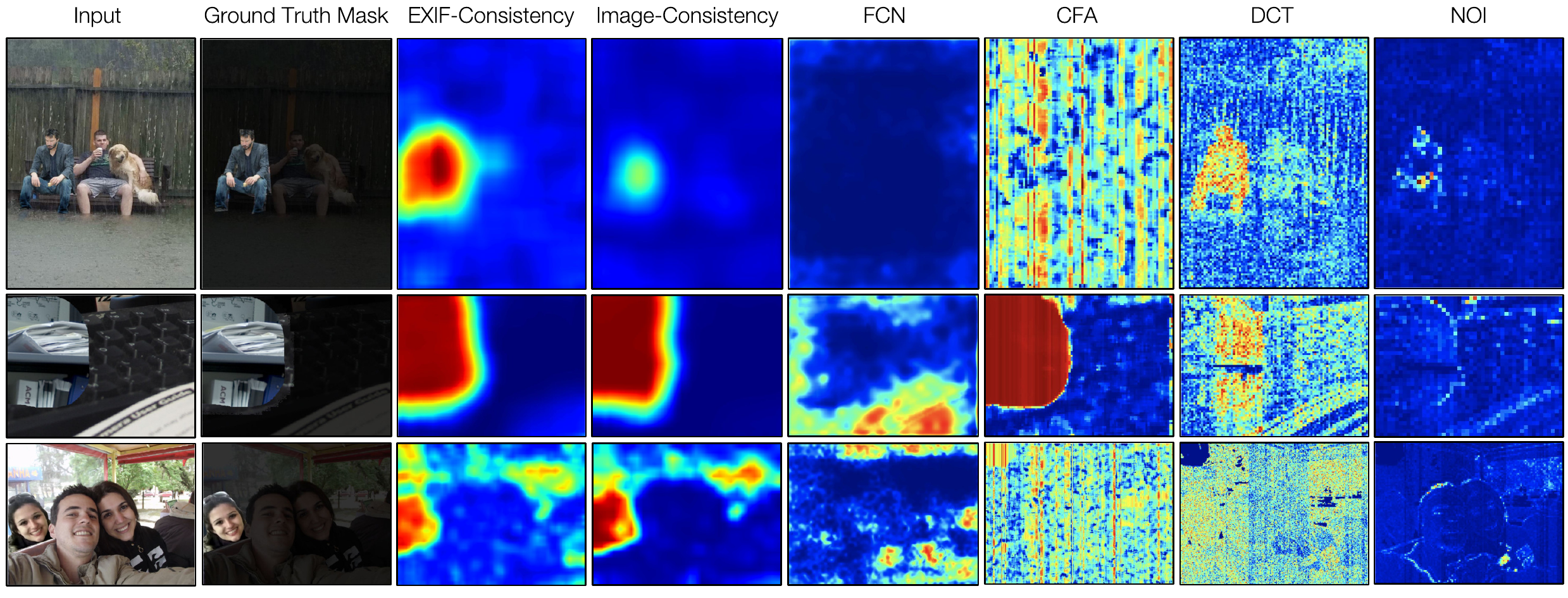}
    \caption{\textbf{Comparing Methods: } We visualize the qualitative difference between \textit{Self-Consistency} and baselines. Our model can correctly localizes image splices from  \textit{In-the-Wild}, \textit{Columbia} and \textit{Carvalho} that other methods make mistakes on.} \vspace{-2mm}
    \label{fig:compare}
\end{figure*}
 
\vspace{-0.5mm}
\subsection{Splice Localization}
\vspace{-1mm}
Having seen that our model can distinguish spliced and authentic images, we next ask whether it can also localize spliced regions within images. For each image, our algorithm produces an unnormalized probability that each pixel is part of a splice.

Because our consistency predictions are relative, it is ambiguous which of the two segments is spliced. We therefore identify the spliced region using a simple heuristic: we say that the smaller of the two consistent regions is the splice. We also consider an alternative evaluation metric that flips (i.e. negates) the consistency predictions if this results in higher accuracy. This measures a model's ability to segment the two regions, rather than its ability to say which is which. In both cases, we evaluate the quality of the localization using mean average precision (mAP). 

We also propose using a per-class intersection over union (cIOU) which averages the IOU of spliced and non-spliced regions after optimal thresholding. 

In order to compare against previous benchmarks ~\cite{salloumfcn}, we also evaluate our results using MCC and F1 measures \footnote{F1 score is defined as {\small $\frac{2TP}{2TP + FN + FP}$} and MCC as {\small $\frac{(TP \times TN) - (FP \times FN)}{\sqrt{(TP + FP)(TP + FN)(TN + FP)(TN + FN)}}$}.}. These metrics evaluate a binary segmentation and require thresholding our predicted probabilities. We use the same evaluation procedure and pick the best threshold per splice localization prediction. Since \cite{salloumfcn} reported their numbers on the full \textit{Columbia} and \textit{Carvalho} datasets (rather than our test split), we evaluated our methods on the full dataset and report the comparison in \tbl{table:splice_usc}.

The quantitative results on \tbl{splice_loc} show that our \textit{EXIF-Consistency} model achieves the best performance across all datasets with the exception of the \textit{Realistic Tampering (RT)} dataset. Notably, the model generally outperformed the supervised baselines, which were trained with actual manipulated images, despite the fact that our model never saw a tampered image during training. The supervised models' poor performance may be due to the small number of artists and manipulations represented in the training data. In \fig{fig:exif-loc}, we show the model's performance on the Columbia dataset when using individual EXIF attributes (rather than the learned ``overall" consistency).

As expected, \textit{EXIF-Consistency} outperformed \textit{Image-Consistency} on most of our evaluations. But, interestingly, we observed that the gap between the models narrowed as training progressed, suggesting that \textit{Image-Consistency} may eventually become competitive with additional training. 

It is also instructive to look at the qualitative results of our method, which we show in \fig{fig:all_results}. 
We see that our method can localize manipulations on a wide range of different splices.  Furthermore, in \fig{fig:neutral-results}, we show that our method produces highly consistent predictions when tested on real images.  We can also look at the qualitative differences between our method and the baselines in \fig{fig:compare}. 

Finally, we ask which EXIF tags were useful for performing the splice localization task. To study this, we computed a response map for individual tags on the Columbia dataset, which we show in \fig{fig:individual}. We see that the most successful tags correspond to imaging parameters that induce photographic changes to the final image like {\tt EXIF DigitalZoomRatio} and {\tt EXIF GainControl}. 

\xpar{Failure cases}
In \fig{fig:failure-results} we show some common failure cases. Our performance on \textit{Realistic Tampering} illustrates some shortcomings with \textit{EXIF-Consistency}. First, our model is not well-suited to finding very small splices, such as the ones that appear in \textit{RT}. When spliced regions are small, the model's large stride may skip over spliced regions, mistakenly suggesting that no manipulations exist.
Second, over- and under-exposed regions are sometimes flagged by our model to be inconsistent because they lack any meta-data signal (\eg because they are nearly uniformly black or white). Finally, \textit{RT} contains a significant number of additional manipulations, such as copy-move, that cannot be consistently detected via meta-data consistency since the manipulated content comes from exactly the same photo.

\xpar{Training and running times}
Training the \textit{EXIF-Consistency} and \textit{Image-Consistency} networks took approximately 4 weeks on 4 GPUs. Running the full self-consistency model took approximately 16 seconds per image (e.g. \fig{fig:compare}). %

\vspace{-2mm}
\section{Discussion}

In this paper, we have proposed a self-supervised method for detecting image manipulations.  Our experiments show that the proposed method obtains state-of-the-art results on several datasets, even though it does not use labeled data during training.  Our work also raises a number of questions. In contrast to physically motivated forensics methods~\cite{farid2016photo}, our model's results are not easily interpretable, and in particular, it is not clear which visual cues it uses to solve the task. It also remains an open question how best to fuse consistency measurements across an image for localizing manipulations. Finally, while our model is trained without any human annotations, it is still affected in complex ways by design decisions that went into the self-supervision task, such as the ways that EXIF tags were balanced during training.

Self-supervised approaches to visual forensics hold the promise of generalizing to a wide range of manipulations --- potentially beyond those that can feasibly be learned through supervised training. However, for a forensics algorithm to be truly general, it must also model the actions of intelligent forgers that adapt to the detection algorithms. Work in adversarial machine learning~\cite{goodfellow2014,szegedy2013intriguing} suggests that having a self-learning forger in the loop will make the forgery detection problem much more difficult to solve, and will require new technical advances. 

As new advances in computer vision and image-editing emerge, there is an increasingly urgent need for effective visual forensics methods. We see our approach, which successfully detects manipulations without seeing examples of manipulated images, as being an initial step toward building general-purpose forensics tools.

\xpar{Acknowledgements} This work was supported, in part, by DARPA MediFor program and UC Berkeley Center for Long-Term Cybersecurity.
We thank Hany Farid and Shruti Agarwal for their advice, assistance, and inspiration in building this project, David Fouhey, Saurabh Gupta, and Allan Jabri for helping with the editing, Peng Zhou for helping with experiments, and Abhinav Gupta for letting us use his GPUs.  
Finally, we thank the many {\em Reddit} and {\em Onion} artists who unknowingly contributed to our dataset.

 {\small
  \bibliographystyle{splncs}
\bibliography{forensics}
}

\newpage
\renewcommand{\thesection}{A\arabic{section}}
\renewcommand{\thefigure}{A\arabic{figure}}
\setcounter{section}{0}
\setcounter{figure}{0}

\section{Appendix}
\label{sec:exif_def}
\xpar{EXIF attribute definitions} We have abbreviated the definitions that were originally sourced from \href{http://www.exiv2.org/tags.html}{http://www.exiv2.org/tags.html}. Please visit our website for additional EXIF information such as: distributions, common values, and prediction rankings.

\scriptsize
\rowcolors{1}{lightgray}{white}
\begin{tabularx}{\textwidth}{p{4.3cm}|p{7.7cm}}
\hline
\multicolumn{1}{c|}{\cellcolor{white}\textbf{EXIF Attribute}} & 
\multicolumn{1}{c}{\cellcolor{white}\textbf{Definition}}\\
\hline
{\tt EXIF BrightnessValue} & The value of brightness.	 \\
{\tt EXIF ColorSpace}	     & The color space information tag is always recorded as the color space specifier. Normally sRGB is used to define the color space based on the PC monitor conditions and environment. If a color space other than sRGB is used, Uncalibrated is set. Image data recorded as Uncalibrated can be treated as sRGB when it is converted to FlashPix. \\
{\tt EXIF ComponentsConfiguration} & Information specific to compressed data. The channels of each component are arranged in order from the 1st component to the 4th. For uncompressed data the data arrangement is given in the tag. However, since can only express the order of Y, Cb and Cr, this tag is provided for cases when compressed data uses components other than Y, Cb, and Cr and to enable support of other sequences.\\
{\tt EXIF CompressedBitsPerPixel} & Specific to compressed data; states the compressed bits per pixel.\\
{\tt EXIF Contrast} & This tag indicates the direction of contrast processing applied by the camera when the image was shot.\\
{\tt EXIF CustomRendered}	 & This tag indicates the use of special processing on image data, such as rendering geared to output. When special processing is performed, the reader is expected to disable or minimize any further processing.\\
{\tt EXIF DateTimeDigitized}	& The date and time when the image was stored as digital data. \\
{\tt EXIF DateTimeOriginal} & The date and time when the original image data was generated. \\
{\tt EXIF DigitalZoomRatio}	& This tag indicates the digital zoom ratio when the image was shot. If the numerator of the recorded value is 0, this indicates that digital zoom was not used.\\
{\tt EXIF ExifImageLength} & The number of rows of image data. In JPEG compressed data a JPEG marker is used instead of this tag.\\
{\tt EXIF ExifImageWidth}	& The number of columns of image data, equal to the number of pixels per row. In JPEG compressed data a JPEG marker is used instead of this tag.\\
{\tt EXIF ExifVersion} & The version of this standard supported. Nonexistence of this field is taken to mean nonconformance to the standard\\
{\tt EXIF ExposureBiasValue} & The exposure bias.	 \\
{\tt EXIF ExposureMode} & This tag indicates the exposure mode set when the image was shot. In auto-bracketing mode, the camera shoots a series of frames of the same scene at different exposure settings.\\
{\tt EXIF ExposureProgram} & The class of the program used by the camera to set exposure when the picture is taken.\\
{\tt EXIF ExposureTime} & Exposure time, given in seconds. \\
{\tt EXIF FileSource}	& Indicates the image source. If a DSC recorded the image, this tag value of this tag always be set to 3, indicating that the image was recorded on a DSC.\\
{\tt EXIF Flash} & Indicates the status of flash when the image was shot.	\\
{\tt EXIF FlashPixVersion} & The FlashPix format version supported by a FPXR file.	\\
{\tt EXIF FNumber}	 & The F number.	\\
{\tt EXIF FocalLength}	& The actual focal length of the lens, in mm.	\\
{\tt EXIF FocalLengthIn35mmFilm}	& This tag indicates the equivalent focal length assuming a 35mm film camera, in mm. A value of 0 means the focal length is unknown. Note that this tag differs from the tag. \\
{\tt EXIF FocalPlaneResolutionUnit}	& Unit of measurement for FocalPlaneXResolution and FocalPlaneYResolution.\\
{\tt EXIF FocalPlaneXResolution}	& Number of pixels per FocalPlaneResolutionUnit in ImageWidth direction for main image.	\\
{\tt EXIF FocalPlaneYResolution}	& Number of pixels per FocalPlaneResolutionUnit in ImageLength direction for main image.	\\
{\tt EXIF GainControl}	& This tag indicates the degree of overall image gain adjustment.	\\
{\tt EXIF InteroperabilityOffset}	& Unknown	\\
{\tt EXIF ISOSpeedRatings}	& Indicates the ISO Speed and ISO Latitude of the camera or input device as specified in ISO 12232.	\\
{\tt EXIF LensMake}	& This tag records the lens manufacturer as an ASCII string.	\\
{\tt EXIF LensModel}	& This tag records the lens's model name and model number as an ASCII string.	\\
{\tt EXIF LensSpecification}	& This tag notes minimum focal length, maximum focal length, minimum F number in the minimum focal length, and minimum F number in the maximum focal length, which are specification information for the lens that was used in photography. When the minimum F number is unknown, the notation is 0/0	\\
{\tt EXIF LightSource}	& The kind of light source.	\\
{\tt EXIF MaxApertureValue}	& The smallest F number of the lens. \\
{\tt EXIF MeteringMode}	& The metering mode.	\\
{\tt EXIF OffsetSchema}	& Unknown	\\
{\tt EXIF Saturation}	& This tag indicates the direction of saturation processing applied by the camera when the image was shot.	\\
{\tt EXIF SceneCaptureType}	   & This tag indicates the type of scene that was shot. It can also be used to record the mode in which the image was shot. Note that this differs from the tag.	\\
{\tt EXIF SceneType}         & Indicates the type of scene. If a DSC recorded the image, this tag value must always be set to 1, indicating that the image was directly photographed.	\\
{\tt EXIF SensingMethod}	       & Type of image sensor.	\\
{\tt EXIF SensitivityType}	   & The SensitivityType tag indicates which one of the parameters of ISO12232 is the PhotographicSensitivity tag.	\\
{\tt EXIF Sharpness}	           & This tag indicates the direction of sharpness processing applied by the camera when the image was shot.	\\
{\tt EXIF ShutterSpeedValue}	   & Shutter speed.	\\
{\tt EXIF SubjectArea}	       & This tag indicates the location and area of the main subject in the overall scene.	\\
{\tt EXIF SubjectDistanceRange}  & This tag indicates the distance to the subject.	\\
{\tt EXIF SubSecTime}	           & A tag used to record fractions of seconds for the tag.	\\
{\tt EXIF SubSecTimeDigitized}   & A tag used to record fractions of seconds for the tag.	\\
{\tt EXIF SubSecTimeOriginal}	& A tag used to record fractions of seconds for the tag.	\\
{\tt EXIF UserComment}	    & A tag for Exif users to write keywords or comments.	\\
{\tt EXIF WhiteBalance } 	    & This tag indicates the white balance mode set when the image was shot.	\\
{\tt GPS GPSAltitude}	        & Indicates the altitude based on the reference in GPSAltitudeRef. \\
{\tt GPS GPSAltitudeRef}	    & Indicates the altitude used as the reference altitude. 	\\
{\tt GPS GPSDate}         	& A character string recording date and time information relative to UTC (Coordinated Universal Time).\\
{\tt GPS GPSImgDirection}	    & Indicates the direction of the image when it was captured.	\\
{\tt GPS GPSImgDirectionRef}	& Indicates the reference for giving the direction of the image when it is captured. \\
{\tt GPS GPSLatitude}      	& Indicates the latitude. \\
{\tt GPS GPSLatitudeRef}	    & Indicates whether the latitude is north or south latitude. \\
{\tt GPS GPSLongitude}	    & Indicates the longitude.  \\
{\tt GPS GPSLongitudeRef}	    & Indicates whether the longitude is east or west longitude.	\\
{\tt GPS GPSTimeStamp}	    & Indicates the time as UTC (Coordinated Universal Time). \\
{\tt GPS GPSVersionID}	& Indicates the version of GPS. \\
{\tt Image Artist}	& This tag records the name of the camera owner, photographer or image creator.	\\
{\tt Image Copyright}	& Copyright information.\\
{\tt Image ExifOffset}	& Image ExifOffset. \\
{\tt Image GPSInfo}	& A pointer to the GPS Info IFD.	\\
{\tt Image ImageDescription}	& A character string giving the title of the image. \\
{\tt Image Make}	& The manufacturer of the recording equipment.	\\
{\tt Image Model}	& The model name or model number of the equipment.\\
{\tt Image Orientation}	& The image orientation viewed in terms of rows and columns.	\\
{\tt Image PrintImageMatching}	& Print Image Matching, description needed.	\\
{\tt Image ResolutionUnit}	& The unit for measuring YResolution and XResolution. The same unit is used for both.	\\
{\tt Image Software}	& This tag records the name and version of the software or firmware of the camera or image input device used to generate the image. \\
{\tt Image XResolution}	& Number of pixels per FocalPlaneResolutionUnit in ImageWidth direction for main image.\\
{\tt Image YCbCrPositioning}	& The position of chrominance components in relation to the luminance component.	\\
{\tt Image YResolution}	& Number of pixels per FocalPlaneResolutionUnit in ImageLength direction for main image.	\\
{\tt Inter InteroperabilityIndex}	& Indicates the identification of the Interoperability rule.	\\
{\tt Inter InteroperabilityVersion}	& Interoperability version.	\\
{\tt Inter RelatedImageLength}	& Image height.	\\
{\tt Inter RelatedImageWidth}	& Image width.	\\
\bottomrule
\end{tabularx}

\end{document}